\pdfoutput=1
\documentclass[11pt]{article}

\usepackage[]{EMNLP2022}
\usepackage{danudefs}

\usepackage{times}
\usepackage{latexsym}
\usepackage[T1]{fontenc}
\usepackage[utf8]{inputenc}
\usepackage{microtype}
\usepackage{inconsolata}
\usepackage{color}
\usepackage{soul}
\usepackage{booktabs}
\usepackage{url}
\usepackage{amsmath}
\usepackage{adjustbox}
\usepackage{amsthm}
\usepackage{esvect}
\usepackage{multirow}
\usepackage{subcaption}
\usepackage{braket}

\usepackage{float}

\usepackage[english]{babel}
\usepackage{hyperref}
\addto\captionsenglish{%
}
\addto\extrasenglish{%
}

\usepackage{todonotes}

\makeatletter
\if@todonotes@disabled

\else

\fi
\makeatother

\usepackage{xifthen}

\usepackage{arydshln}
\makeatletter
\def\adl@drawiv#1#2#3{%
        \hskip.5\tabcolsep
        \xleaders#3{#2.5\@tempdimb #1{1}#2.5\@tempdimb}%
                #2\z@ plus1fil minus1fil\relax
        \hskip.5\tabcolsep}
\newcommand{\cdashlinelr}[1]{%
  \noalign{\vskip\aboverulesep
          \global\let\@dashdrawstore\adl@draw
          \global\let\adl@draw\adl@drawiv}
  \cdashline{#1}
  \noalign{\global\let\adl@draw\@dashdrawstore
          \vskip\belowrulesep}}
\makeatother
\title{On the Curious Case of $\ell_2$ norm of Sense Embeddings}

\author{Yi Zhou \\
  University of Liverpool \\
  {\tt y.zhou71@liverpool.ac.uk} \\\And
  Danushka Bollegala \\
  University of Liverpool, Amazon \\
  {\tt danushka@liverpool.ac.uk} \\}

\begin{document}
\maketitle
\begin{abstract}
    We show that the $\ell_2$ norm of a static sense embedding encodes information related to the frequency of that sense in the training corpus used to learn the sense embeddings.
    This finding can be seen as an extension of a previously known relationship for word embeddings to sense embeddings.
    Our experimental results show that, in spite of its simplicity, the $\ell_2$ norm of sense embeddings is a surprisingly effective feature for several word sense related tasks such as (a) most frequent sense prediction,
    (b) Word-in-Context (WiC), and 
    (c) Word Sense Disambiguation (WSD).
    In particular, by simply including the $\ell_2$ norm of a sense embedding as a feature in a classifier, we show that we can improve WiC and WSD methods that use static sense embeddings.
\end{abstract}

\section{Introduction}
\label{sec:intro}
    \paragraph{Background:} Given a text corpus, static word embedding learning methods (\citealt{pennington2014glove}, \citealt{mikolov2013efficient}, etc.) learn a \emph{single} vector (aka \emph{embedding}) to represent the meaning of a word in the corpus.
    In contrast, static sense embedding learning methods (\citealt{loureiro2019language}, \citealt{scarlini2020more}, etc.) learn \emph{multiple} embeddings for each word, corresponding to the different senses of that word.
    
    \newcite{Arora-word} proposed a random walk model on the word co-occurrence graph and showed that if word embeddings are uniformly distributed over the unit sphere, the log-frequency of a word in a corpus is proportional to the squared $\ell_2$ norm of the static word embedding, learnt from the corpus.
    \newcite{Hashimoto:TACL:2016} showed that under a simple metric random walk over words where the probability of transitioning from one word to another depends only on the squared Euclidean distance between their embeddings, the log-frequency of word co-occurrences between two words converges to the negative squared Euclidean distance measured between the corresponding word embeddings.
    \newcite{mu2018allbutthetop} later showed that word embeddings are distributed in a narrow cone, hence not satisfying the uniformity assumption used by \newcite{Arora-word}, however their result still holds for such anisotropic embeddings.
    On the other hand, \newcite{Arora:-sense} showed that a word embedding can be represented as the linearly-weighted combination of sense embeddings.
    However, to the best of our knowledge, it remains unknown thus far as to \textbf{What is the relationship between the sense embeddings and the frequency of a sense?}, the central question that we study in this paper.
    
    \paragraph{Contributions:} 
    First, by extending the prior results for word embeddings into sense embeddings, we show that the \textbf{squared $\ell_2$ norm of a static sense embedding is proportional to the log-frequency of the sense in the training corpus.} 
    This finding has important practical implications.
    For example, it is known that assigning every occurrence of an ambiguous word in a corpus to the most frequent sense of that word (popularly known as the Most Frequent Sense (MFS) baseline) is a surprisingly strong baseline for WSD~\cite{mccarthy2004finding, McCarthy:2007}. 
    Therefore, the theoretical relationship which we prove implies that we should be able to use $\ell_2$ norm to predict the MFS of a word.
        
    Second, we conduct a series of experiments to empirically validate the above-mentioned relationship. 
    We find that the relationship holds for different types of static sense embeddings learnt using methods such as GloVe~\cite{pennington2014glove} and skip-gram with negative sampling~\cite[SGNS;][]{mikolov2013distributed} on SemCor~\cite{miller1993semantic}.
    
    Third, motivated by our finding that $\ell_2$ norm of pretrained static sense embeddings encode sense-frequency related information,
    we use $\ell_2$ norm of sense embeddings as a feature for several sense-related tasks such as (a) to predict the MFS of an ambiguous word, (b) determining whether the same sense of a word has been used in two different contexts~\cite[WiC;][]{Pilehvar:2019}, and (c) disambiguating the sense of a word in a sentence (WSD).
    We find that, regardless of its simplicity, $\ell_2$ norm is a surprisingly effective feature, consistently improving the performance in all those benchmarks/tasks.
    % The evaluation scripts have been submitted anonymously via the submission system and will be publicly released upon paper acceptance.
    The evaluation scripts is available at: \url{https://github.com/LivNLP/L2norm-of-sense-embeddings}.

% Tell these things somewhere in the paper.
% We do not do WiC. This will mean we will just get 50% as all distinct (negatively labelled) sentence pairs will always be predicted as having the same sense (as positive) by a method that does sense prediction based on MFS.
% We can do WSD but that would only show how well the MFS baseline can be used.
% We can say that l2 norm is a strong feature in its own that can be helpful for real world WSD systems. Now, l2 norm might be already being used in some existing WSD systems (check papers! and cite) but we, for the first-time, show that this feature is theoretically guaranteed to be an indicator of sense frequency.

\section{$\ell_2$ norm vs. Frequency}
\label{sec:theory}
% This is the theoretical section. DB will try to prove this. 
% We will put the proof sketch here and defer it to supplementary, if too long. (max 0.5 page)

Let us first revisit the generative model proposed by \newcite{Arora-word} for static word embeddings, where the $t$-th word, $v$, in a corpus is generated at step $t$ of a random walk of a \emph{context} vector $\vec{c}_t$, which represents what is being talked about.
The probability, $p(v|c_t)$, of emitting $v$ at time $t$ is modelled using a log-linear word production model, proportionally to $\exp(\vec{c}_t\T\vec{v})$.
If $\cG_v$ is a \emph{word co-occurrence} graph, where vertices correspond to the words in the vocabulary, $\cV$, the random walker can be seen as visiting the vertices in $\cG_v$ according to this probability distribution.
\newcite{Arora-word} showed that the partition function, $Z_c$, given by \eqref{eq:partition} for this probabilistic model is a constant $Z$, independent of the context $c$.% which is known as \emph{self-normalising}~\cite{andreas-klein:2015:NAACL-HLT}.
\begin{align}
\label{eq:partition}
Z_c = \sum_v \exp(\vec{c}\T\vec{v})
\end{align}
Assuming that the stationary distribution of this random walk is uniform over the unit sphere, \newcite{Arora-word} proved the relationship in \eqref{eq:arora-word}, for $d$ dimensional word embeddings, $\vec{v} \in \R^d$.
\begin{align}
    \label{eq:arora-word}
    \log p(v) = \frac{\norm{\vec{v}}_2^2}{2d} - \log Z 
\end{align}
Let the frequency of $v$ in the corpus be $f(v)$, and the total number of word occurrences be $N = \sum_{v} f(v)$.
$p(v)$ can be estimated using corpus counts as $f(v)/N$.
Because $N$, $d$, and $Z$ are constants, independent of $v$, \eqref{eq:arora-word} implies a linear relationship between $\log f(v)$ and $\norm{\vec{v}}_2^2$.

% we will later show self-normalisation is true even for sense embeddings.
% show the word embedding result for such a random walk by Arora
% Now explain the extended random walk over the sense co-occurrence graph

\begin{figure}[t]
\centering
\includegraphics[width=0.48\textwidth,height=0.26\textwidth]{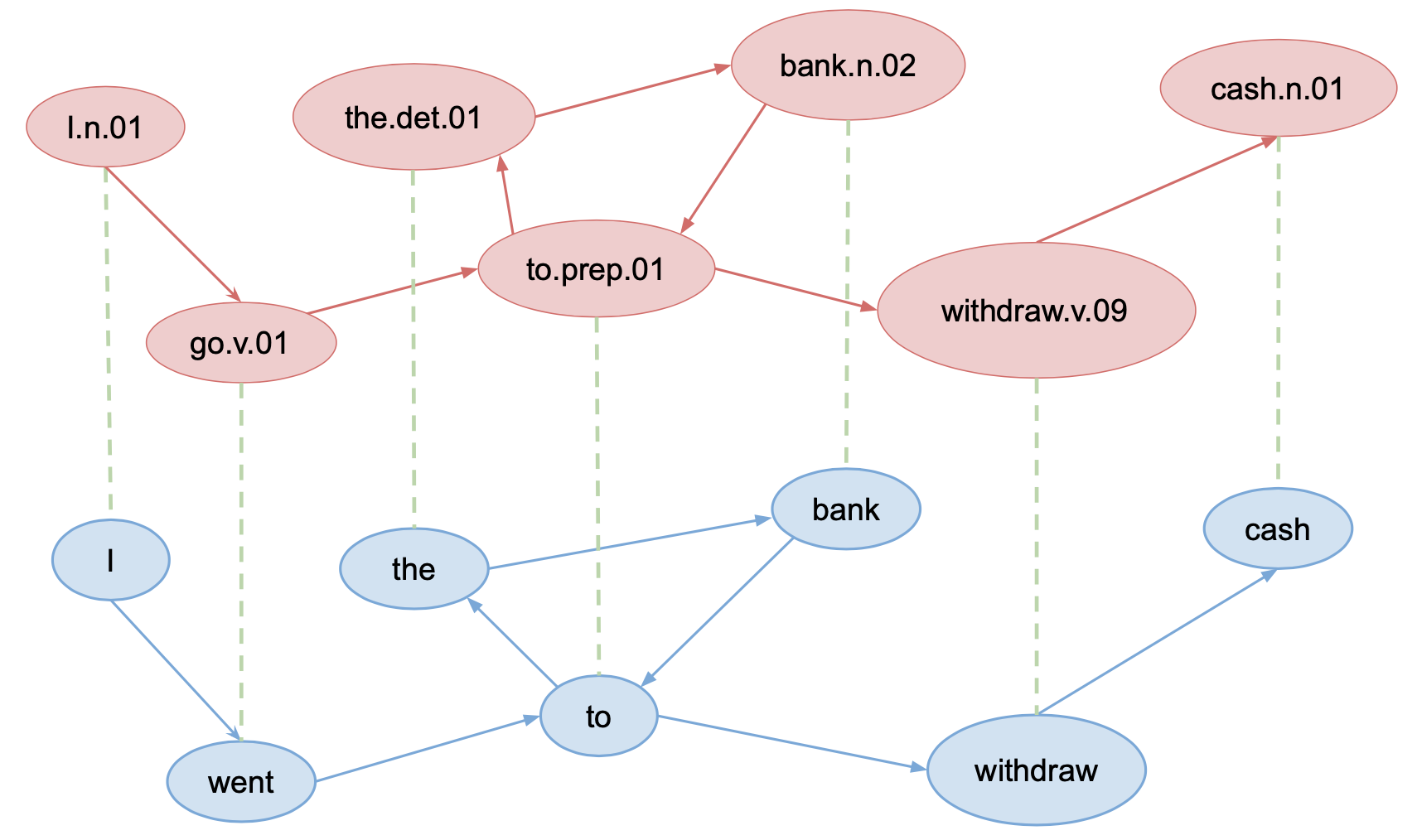}
\caption{Part of the word co-occurrence graph $\cG_v$ (bottom) shown with the corresponding sense co-occurrence graph $\cG_s$ (top). Each word in $\cG_v$ is mapped to its correct sense in $\cG_s$.}
\label{fig:random-walk}
% \vspace{-3mm}
\end{figure}

To extend this result to sense embeddings, we observe that the word $v$ generated at step $t$ by the above-described random walk can be uniquely associated with a sense id $s_v$, corresponding to the meaning of $v$ as used in $c_t$.
If we consider a second \emph{sense co-occurrence} graph $\cG_s$, where vertices correspond to the sense ids, then the above-mentioned corpus generation process corresponds to a second random walk on $\cG_s$, as shown in~\autoref{fig:random-walk}.

Although an ambiguous word can be mapped to multiple sense ids across the corpus in different contexts, at any given time step $t$, a word $v$ is mapped to only one vertex in $\cG_s$, determined by the context $c_t$.
Indeed a WSD can be seen as the process of finding such a mapping.
The two random walks over word and sense id graphs are isomorphic and converge to the same set of final states~\cite{Bauerschmidt_2021}.
Therefore, an analogous relationship given by \eqref{eq:arora-sense} can be obtained by replacing word embeddings, $\vec{v}$, with sense embeddings, $\vec{s}$, in \eqref{eq:arora-word}.
\begin{align}
    \label{eq:arora-sense}
    \log p(s) =  \frac{\norm{\vec{s}}_2^2}{2d_s} - \log Z'
\end{align}
Here, $d_s$ is the dimensionality of the sense embeddings $\vec{s} \in \R^{d_s}$.
Later in \autoref{sec:exp}, we empirically show that the normalisation coefficient, $Z' = \sum_s \exp(\vec{c}\T\vec{s})$, for sense embeddings also satisfies the self-normalising~\cite{andreas-klein:2015:NAACL-HLT} property, thus independent of $c$.
If we abuse the notation $f(s)$ to denote also the frequency of $s$ in the corpus (i.e. the total number of times the random walker visits the vertex $s$), from \eqref{eq:arora-sense} it follows that $\log f(s)$ is linearly related to $\norm{\vec{s}}_2^2$.

\section{Empirical Validation}
\label{sec:exp}

% 1 page
% Jodie TODO:
% Explain the static sense embedding learning method using GloVe, SGNS on SemCor.
% Explain SemCor a bit if necessary
% Show figures for l2 norm and frequency for Glove (or SGNS. we can keep one of these figures in the main body and move the rest to sup)
% Show that this does not hold for LMMS and tell it is because it is based on BERT, which is not a static word embedding and hence the random walk model we used in the theoretical proof does not hold.
% we might have to argue that static word embeddings are still useful as in this paper \cite{arora-etal-2020-contextual}
% This could be a weak point as our result does not hold for sota sense embeddings. We have to explain this carefully.

The theoretical analysis described in \autoref{sec:theory} implies a linear relationship between $\log f(s)$ and $\norm{\vec{s}}_2^2$ for the learnt sense embeddings.
To empirically verify this relationship, we learn static sense embeddings using GloVe and SGNS from SemCor, which is the largest corpus manually annotated with WordNet~\cite{miller1995wordnet} sense ids.
Specifically, we consider the co-occurrences of senses instead of words for this purpose.
To distinguish the sense embeddings learnt from GloVe and SGNS from their word embeddings, we denote these by respectively \textbf{GloVe-sense} and \textbf{SGNS-sense}.

\begin{figure}[t]
\centering
\includegraphics[width=0.48\textwidth,height=0.35\textwidth]{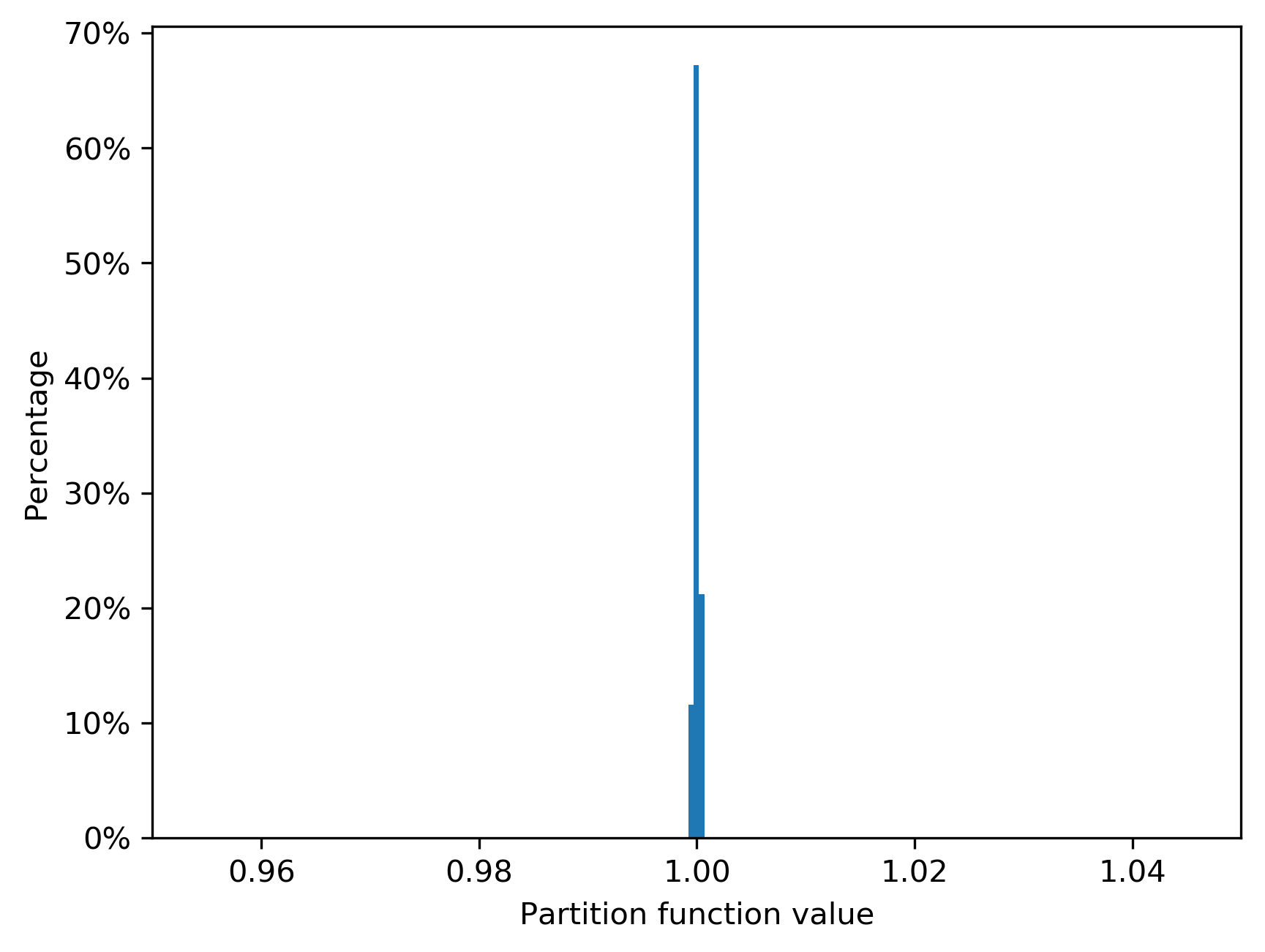}
\caption{Histogram of the partition function for 1,000 random vectors $\vec{c}$ for GloVe-sense. The $x$-axis is normalised by the mean of the values.}
\label{fig:glove-partition}
%\vspace{-3mm}
\end{figure}

\begin{figure}[t]
\centering
\includegraphics[width=0.48\textwidth,height=0.35\textwidth]{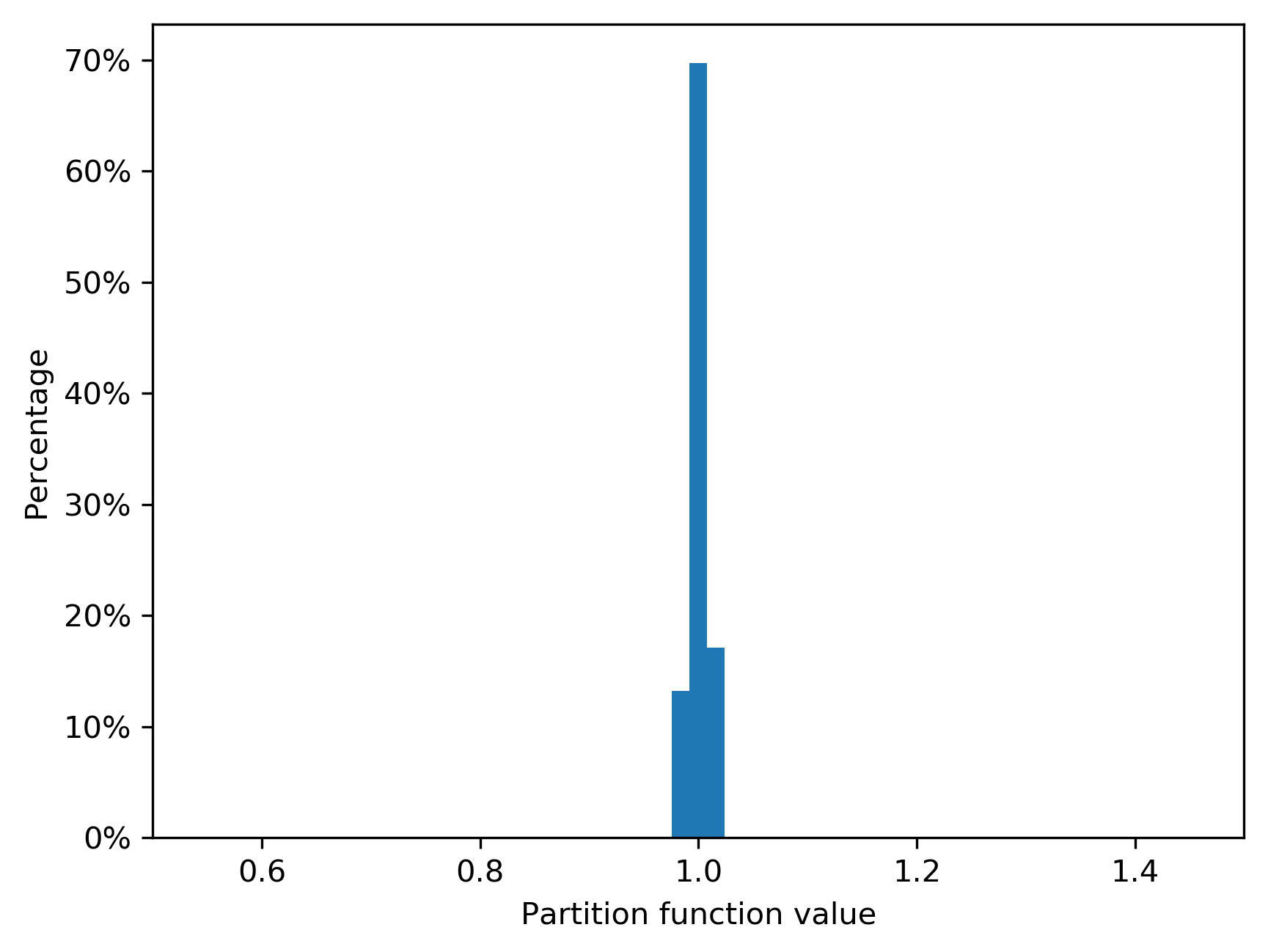}
\caption{Histogram of partition function for 1,000 random vectors $\vec{c}$ for SGNS-sense. The $x$-axis is normalised by the mean of the values.}
\label{fig:sgns-partition}
%\vspace{-3mm}
\end{figure}

\autoref{fig:glove-partition} shows the partition function for GloVe-sense embeddings.
We see that the partition function is tightly concentrated around its mean, showing that sense-embeddings also demonstrate self-normalisation similar to word embeddings.
% The partition function for SGNS-sense shown in \autoref{fig:sgns-partition} shows a similar distribution.
% \autoref{fig:sgns-partition} shows the the partition function for the SGNS-sense embeddings. 
Similar to the histogram for GloVe-sense embeddings, we see that the partition function for SGNS-sense embeddings is also tightly centred around the mean (i.e., $1.0$) from \autoref{fig:sgns-partition}. 

\begin{figure}[t]
\centering
\includegraphics[width=0.48\textwidth,height=0.35\textwidth]{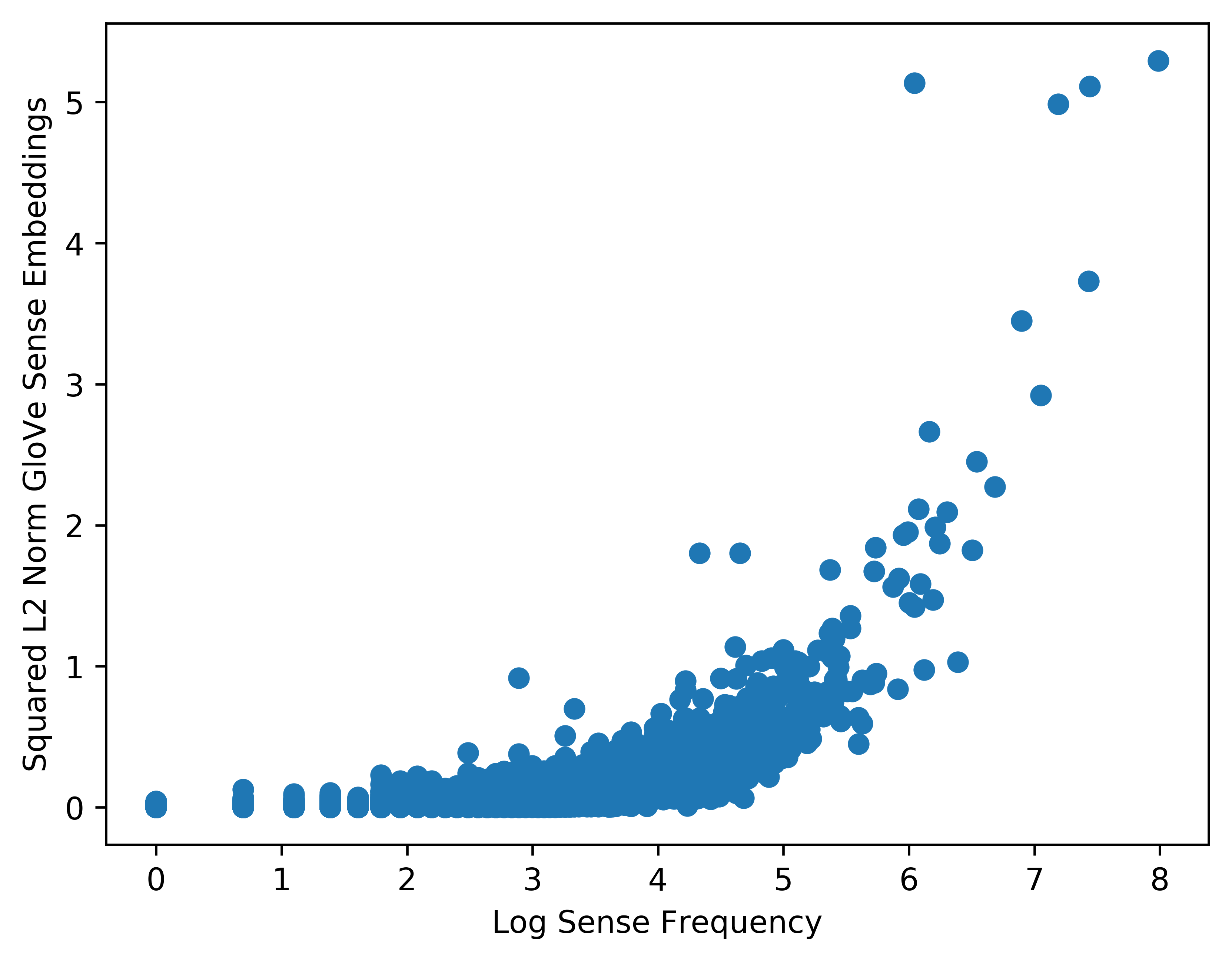}
\caption{A linear relationship between $\log f(s)$ ($x$-axis) and $\norm{\vec{s}}_2^2$ ($y$-axis) can be seen for GloVe-sense embeddings represented by the blue dots.}
\label{fig:glove-corr}
%\vspace{-4mm}
\end{figure}

\begin{figure}[t]
\centering
\includegraphics[width=0.48\textwidth,height=0.35\textwidth]{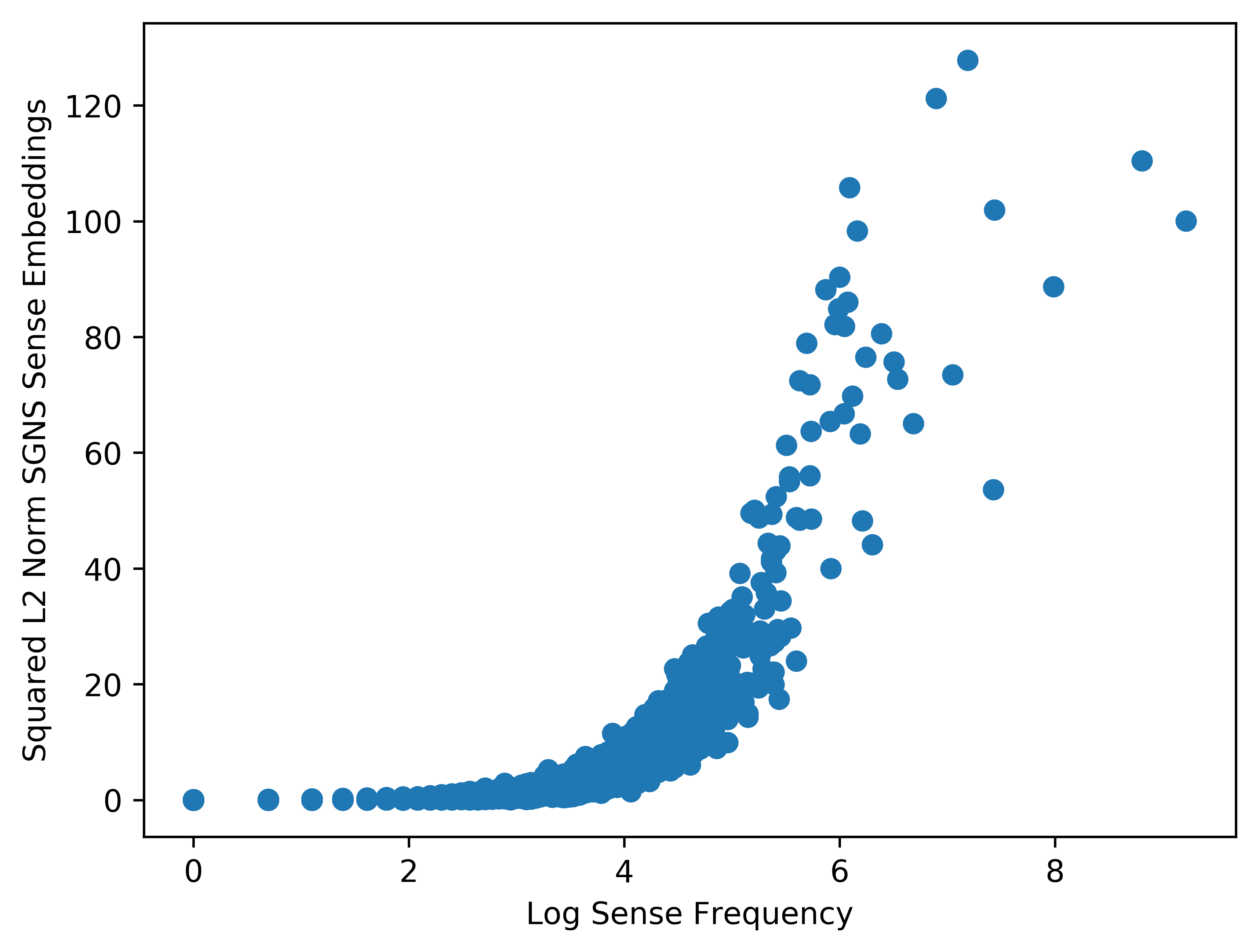}
% \caption{A linear relationship can be seen between the squared $\ell_2$ norms of SGNS-sense embeddings and the logarithms of the sense frequencies. Each dot in the plot corresponds to a sense, where x-axis is the natural logarithm of the sense frequency, and y-axis is the squared $\ell_2$ norm of the word embedding. The Pearson correlation coefficient between the two variables is $0.440$.}
\caption{A linear relationship between $\log f(s)$ ($x$-axis) and $\norm{\vec{s}}_2^2$ ($y$-axis) can be seen for SGNS-sense embeddings represented by the blue dots.}
\label{fig:sgns-corr}
%\vspace{-2.8mm}
\end{figure}

\autoref{fig:glove-corr} shows the correlation between $\log f(s)$ and $\norm{\vec{s}}_2^2$ for GloVe-sense.
We see a moderate positive correlation (Pearson's $\rho = 0.437$) between these two variables, confirming the linear relationship predicted in \autoref{sec:theory}.
% Correlation plot for SGNS-sense (in Appendix) shows a similar linear relationship.
% \autoref{fig:sgns-corr} shows a linear relationship between the squared $\ell_2$ norm of SGNS-sense embeddings and the logarithm of the corresponding sense frequencies. 
Similar to the correlation plot for GloVe-sense embeddings, from \autoref{fig:sgns-corr}, one can see a positive correlation (Pearson's $\rho = 0.440$) between the log-frequency and squared $\ell_2$ norm for the SGNS-sense embeddings. 

\begin{table}[t]
\centering
\resizebox{0.48\textwidth}{!}{
\begin{tabular}{lcc}
\toprule
Models & All words  & Noun Sample \\
\midrule
Random & 67.6 & 26.0 \\
% MKWC04~\cite{mccarthy2004finding} & n/a & 54 \\
UMFS-WE & 73.9 & 48.0 \\ %\cite{bhingardive2015unsupervised}
EnDi & 71.4 & 47.4 \\ %~\cite{pasini2018two}
WCT-VEC & 75.2 & 48.8 \\ %~\cite{hauer2019you}
COMP2SENSE & 77.9 & 58.5 \\ %~\cite{hauer2019you}
\midrule
\textit{Ours} \\
GloVe-sense with $\ell_2$ norm & 90.1 & 92.2 \\
SGNS-sense with $\ell_2$ norm & \pmb{95.6} & \pmb{96.6} \\
\bottomrule
\end{tabular}}
\caption{Percentage accuracy for the MFS prediction task on SemCor for All Words and the Noun Sample, limited to polysemous nouns. Overall best scores are in bold.}
\label{tbl:mfs}
\vspace{-1.8mm}
\end{table}

It is noteworthy however this linear relationship between log-frequency and squared $\ell_2$ norm does not hold for contextualised word embeddings such as BERT~\cite{devlin2019bert} or static sense embeddings such as LMMS~\cite{loureiro2019language} that are computed by averaging BERT embeddings (see  \autoref{sec:context} for details).
The random walk model described in \autoref{sec:theory} cannot be applied to contextualised embeddings because the probability of occurrence of a word under the discriminative masked language modelling objectives used to train contextualised word embeddings such as BERT depend on all the words generated before as well as after the target word.
%Due to this reason the frequency of an LMMS sense embedding is not linearly related to the squared $\ell_2$ norm of the sense embedding.

\subsection{Predicting the Most Frequent Sense}
% 1 page
% We explain the task, dataset and baseline methods.
% We need to be clear that as we apply our method it is not unsupervised because we do need a sense-tagged corpus (SemCor) to learn the sense embeddings using GloVe here.
% Show results for All words and Nouns only. No need to talk about other norms as they are theoretically invalid.

% We can also show results for the MFS vs. NFS prediction task (which is more difficult than the previous task)

To investigate whether frequency of a sense is indeed represented by the squared $\ell_2$ norm of its static sense embedding, we conduct an MFS prediction task on SemCor following the setup proposed by~\newcite{hauer2019you}.
In this MFS prediction task, given the set of senses of an ambiguous word, we must predict the sense with the highest frequency for that word in  SemCor.
For this purpose, we filter senses by the lemma and part-of-speech (PoS) of the target word and select the sense with the largest squared $\ell_2$ norm using GloVe-sense and SGNS-sense embeddings separately.

%As the proposed method is not unsupervised, we use our pre-trained GloVe-sense and SGNS-sense embeddings for this task.
%We predict the MFS by means of the $\ell_2$ norm of sense embeddings.
%Specially, for an ambiguous word, we first obtain a set containing all of its candidate senses based on its lemma and part-of-speech (pos). 
%Then we compute the $\ell_2$ norm of static sense embeddings for each sense from the candidate set.
%Finally, we select the sense with the highest $\ell_2$ norm as the prediction. 

In \autoref{tbl:mfs}, we compare our results against a random sense selection baseline and several prior proposals on the MFS benchmark dataset~\cite{hauer2019you}.
EnDi~\cite{pasini2018two} is a language-independent and fully automatic method for sense distribution learning from raw text.
%We also compare our method against three other approaches that identify the MFS of a word, namely,
UMFS-WE~\cite{bhingardive2015unsupervised} and WCT-VEC~\cite{hauer2019you} both use the distance between word and sense embeddings.
COMP2SENSE~\cite{hauer2019you} is a knowledge-based method using WordNet and uses a set of words known as the companions of a target word to determine MFS, based on a sense-similarity function.
As seen from \autoref{tbl:mfs}, both GloVe-sense and SGNS-sense outperform all the other methods for \textbf{all} words and \textbf{noun} sample settings.
In particular, for \textbf{noun} sample, which contains polysemous nouns that occur at least 3 times in SemCor, both methods obtain more than $35\%$ accuracy improvements over the next best method, providing strong empirical evidence supporting the linear relationship predicted by \eqref{eq:arora-sense}.
%This result provides strong support to our hypothesis that $\ell_2$ norm of static sense embeddings convey a strong signal about the word’s MFS.

If the $\ell_2$ norm of a sense embedding relates to the frequency of that sense, the $\ell_2$ norm of the most frequent sense should be always greater than the $\ell_2$ norm of the next frequent sense of an ambiguous word.
To further investigate this, we sort the set of ambiguous words in SemCor based on their frequency and divide them into 10 subsets (i.e., bins).
The summary statistics of each subset is shown in \autoref{tbl:bins}.
% To further study whether $\ell_2$ norms of sense embeddings encode frequency information of the corresponding ambiguous words, we sort the set of ambiguous words in SemCor based on their frequency and divide them into 10 subsets (i.e., bins).
For each ambiguous word $w$, we find its most frequent sense $w_m$ and next frequent sense $w_n$ in SemCor. 
Note that $w_m$ and $w_n$ are determined based on their frequency in SemCor and not according to how they are sorted in the WordNet.

\begin{table}[t]
\centering
\small
\begin{tabular}{c c c c}
\toprule 
Bins & Max Freq & Min Freq & Word Count \\
    % & Frequency &  Frequency &  Count \\
\midrule        
1    & 15,783  & 64    & 545 \\
2    & 64      & 32    & 545 \\
3    & 32      & 20    & 545 \\
4    & 20      & 13    & 545 \\
5    & 13     & 9   & 545 \\
6    & 9     & 7   & 545 \\
7    & 7     & 5   & 545 \\
8    & 5     & 3   & 545 \\
9    & 3     & 2   & 545 \\
10    & 2    & 1   & 545 \\
\bottomrule
\end{tabular}
\caption{Statistics of each bin of ambiguous words grouped based on their frequency in SemCor.}
\label{tbl:bins}
% \vspace{-0.65mm}
\end{table}

\begin{figure}[t]
\centering
\includegraphics[width=0.48\textwidth,height=0.35\textwidth]{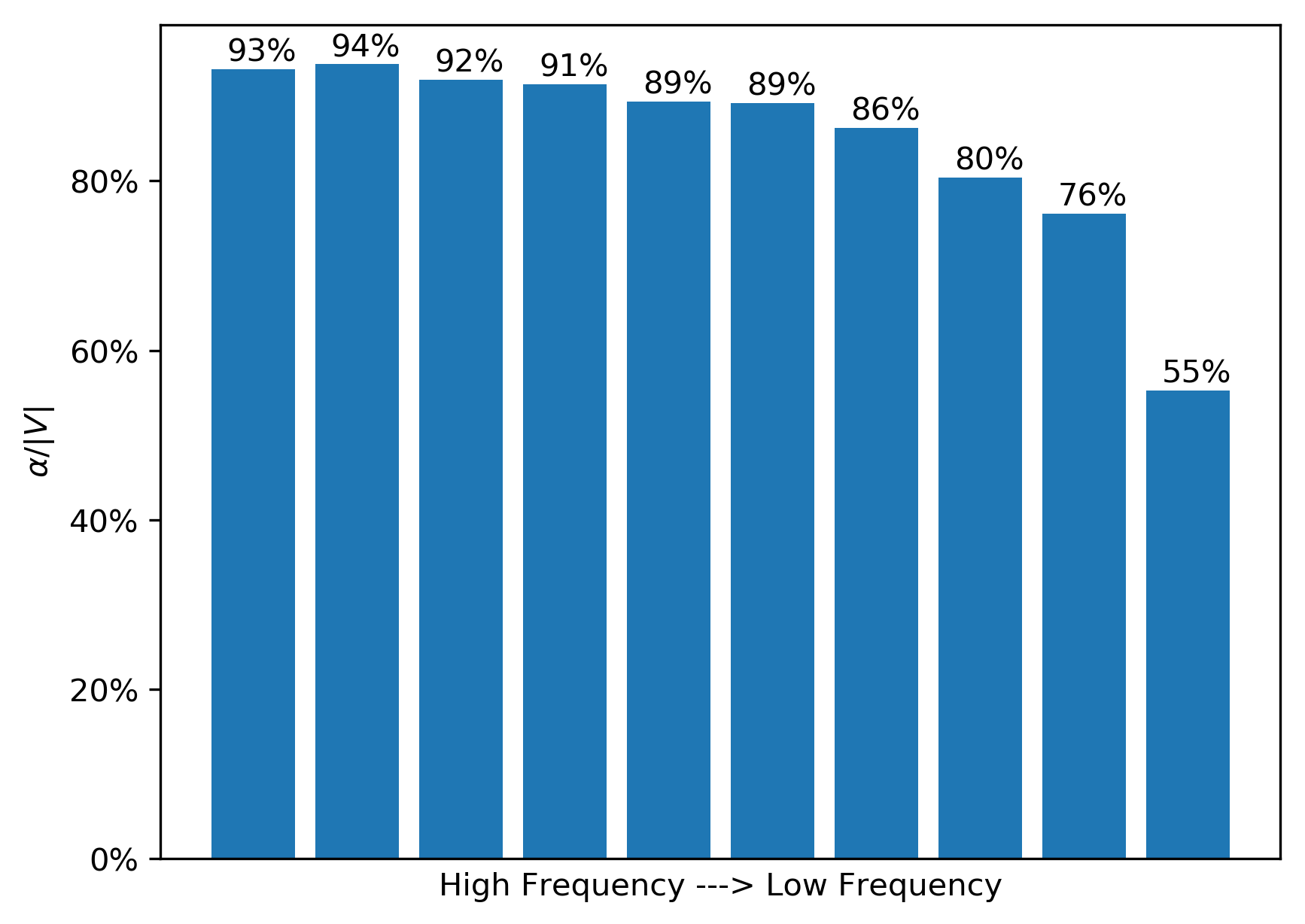}
\caption{The trend of $\alpha/|\mathcal{V}|$ from high frequent words to low frequent words.}
\label{fig:mfs-nfs}
\end{figure}

Let us denote the $\ell_2$ norms of $w_m$ and $w_n$ by $\norm{\vec{w_m}}_2$ and $\norm{\vec{w_n}}_2$ respectively, and $\alpha = \sum_{w\in\mathcal{V}}\mathbb{I}(\textbf{$\norm{\vec{w_m}}_2$ > $\norm{\vec{w_n}}_2$})$, where $\mathbb{I}(x)$ is the indicator function that returns 1 if $x$ is True and 0 otherwise. 
$\mathcal{V}$ is the set of ambiguous words (i.e., the words that have a least two distinct senses in SemCor).
We compute the percentage $\alpha/|\mathcal{V}|$ for the total vocabulary and shown the result in \autoref{fig:mfs-nfs}.
We observe that the second bin obtains the highest $\alpha/|\mathcal{V}|$ score over all the bins. 
Moreover, the $\alpha/|\mathcal{V}|$ scores decrease with the frequency of the ambiguous words. 
This result indicates that the relationship between the $\ell_2$ norm of a sense embedding and its frequency is stronger for high frequent words than low frequent ones.

\subsection{Predicting Word Sense in Context}
\label{sec:WiC}

% \begin{table}[t]
% \centering
% \small
% \begin{tabular}{lcccc}
% \toprule
% Models &Dev &Test  \\
% \midrule
% \textit{LMMS-based} \\
% LMMS~\cite{loureiro2019language} & 69.4 & 64.8 \\
% LMMS + $\ell_2$ norm of GloVe-sense & \pmb{71.3} & 65.8 \\
% LMMS+$\ell_2$ norm of SGNS-sense  & 70.5 & \pmb{67.0} \\
% \midrule
% \textit{ARES-based} \\
% ARES~\cite{scarlini2020more} & 67.6 & 66.6  \\
% ARES+$\ell_2$ GloVe-sense & 67.6 & 66.6 \\
% ARES+$\ell_2$ SGNS-sense & 67.6 & 66.7 \\
% \bottomrule
% \end{tabular}
% \caption{Accuracies on the WiC dev and test sets for LMMS- (top) and ARES- (bottom) based classifiers.}
% \label{tbl:wic}
% \vspace{-3mm}
% \end{table}

\begin{table}[t]
\centering
\begin{tabular}{lcccc}
\toprule
Models &Test  \\
\midrule
\textit{LMMS-based} \\
LMMS~\cite{loureiro2019language} & 64.8 \\
LMMS + $\ell_2$ norm of GloVe-sense & 65.8 \\
LMMS+$\ell_2$ norm of SGNS-sense  & \pmb{67.0} \\
\midrule
\textit{ARES-based} \\
ARES~\cite{scarlini2020more}  & 66.6  \\
ARES+$\ell_2$ GloVe-sense  & 66.6 \\
ARES+$\ell_2$ SGNS-sense  & 66.7 \\
\bottomrule
\end{tabular}
\caption{Accuracies on the WiC dev and test sets for LMMS- (top) and ARES- (bottom) based classifiers. Overall best score is in bold.}
\label{tbl:wic}
\vspace{-2mm}
\end{table}

We evaluate the $\ell_2$ norm of sense embeddings in WiC and WSD as downstream tasks.
In WiC, given an ambiguous word $w$ occurring in two contexts $c_1$ and $c_2$, we must predict whether $w$ occurs in $c_1$ and $c_2$ with the same sense.
We follow~\newcite{loureiro2019liaad}, and train a binary logistic regression model on WiC training set using different sets of similarities between static sense embeddings and contextualised embeddings obtained from a language model (i.e. BERT) as features. Further details regarding the features used for this classifier are given in \autoref{appendix:sec:WSD}.
We consider two current state-of-the-art sense embeddings, LMMS and ARES~\cite{scarlini2020more}, and include $\ell_2$ norm of static sense embeddings as extra features, and measure the gain in performance.

From \autoref{tbl:wic} we see that by including $\ell_2$ norm of GloVe-sense and SGNS-sense embeddings as features, we obtain more than $1\%$ gains in accuracy over the original LMMS on WiC dev and test sets.
ARES+$\ell_2$ norm GloVe-sense obtains the same score as the ARES baseline, while ARES+$\ell_2$ norm SGNS-sense achieves a slight improvement on the test set.  
This shows that $\ell_2$ norm of static sense embeddings encodes sense frequency related information, which improves the performance in WiC when used with static sense embeddings.
This is noteworthy given that $\ell_2$ norm is a single feature compared to LMMS and ARES, which are both 2048 dimensional.

\subsection{Word Sense Disambiguation}
\label{sec:WSD}

We further evaluate $\ell_2$ norm of static sense embeddings using the English all-words WSD framework~\cite{raganato2017word}.
%To avoid any discrepancies in the scoring methodology, we use the framework's scoring scripts. 
For this purpose, we train a binary logistic regression classifier using the two features -- (a) the similarity between the contextualised embedding and a sense embedding of the target word, and (b) the squared $\ell_2$ norm of the sense embedding. 
We use SemCor training data and consider the correct sense of the target word as a positive instance, and its other senses as negative instances.
At inference time, we predict the sense with the highest probability of being positive as the correct sense of the test word in the given context. 
Likewise in the WiC evaluation in \autoref{sec:WiC}, we measure the improvements in performance over LMMS and ARES, with using $\ell_2$ norm as a feature for WSD. 

\begin{table}[t]
\centering
\resizebox{0.48\textwidth}{!}{
\begin{tabular}{lcccccc}
\toprule
Methods &SE2 &SE3 &SE07 &SE13 &S15 &ALL \\
\midrule
\textit{LMMS-based} \\
LMMS &76.3 &75.6 &68.1 &75.1 &77.0 &75.4 \\
LMMS+$\ell_2$ norm GloVe-sense &77.8 &76.9 &70.5 &76.6 &77.8 &76.8 \\
LMMS+$\ell_2$ norm SGNS-sense &77.5 &77.4 &69.7 &77.1 &78.1 &76.9 \\
\midrule
\textit{ARES-based} \\
ARES &78.0 &77.1 &71.0 &77.3 &\pmb{83.2} &77.9 \\
ARES+$\ell_2$ norm GloVe-sense &\pmb{78.4} &\pmb{77.8} &\pmb{71.6} &77.9 &82.4 &\pmb{78.3} \\
ARES+$\ell_2$ norm SGNS-sense &77.6 &77.5 &68.6 &\pmb{78.0} &82.0 &77.7 \\
\bottomrule
\end{tabular}
}
\caption{F1 on the test sets of the all-words English WSD framework for LMMS- (top) and ARES- (bottom) based method. Overall best scores are in bold.}
\label{tbl:wsd}
\vspace{-1.8mm}
\end{table}

\autoref{tbl:wsd} shows the F1 scores for all-words English WSD datasets.
ARES+$\ell_2$ norm Glove-sense reports the best performance in three out of the five datasets, and obtains the best performance on ALL (i.e., concatenation of all the test sets), whereas  ARES+$\ell_2$ norm SGNS-sense reports the best performance in SE13.
In LMMS-based evaluations, we see that always either one or both GloVE/SGNS-sense $\ell_2$ norms improve over the vanilla LMMS.
This shows that we are able to improve the performance of both LMMS and ARES by simply adding $\ell_2$ norm of static sense embeddings as extra features.

\section{Conclusion}
% If space permits add a related work section. 
% This is not necessary for this paper though if we are explaining various related works (e.g. theoretical works in section 2 and MFS baselines etc. in Section 3,4)
We investigated log-frequency and $\ell_2$ norm of sense embeddings, and found a linear relationship between those.
Experimental results indicate that, despite its simplicity, $\ell_2$ norm of sense embedding is a surprisingly effective feature for MFS prediction, WiC and WSD tasks. 

\section{Limitations}
This paper makes both theoretical and empirical contributions related to sense embeddings.
In this section, we highlight some of the important limitations in terms of both theory and empirical evaluations we made in the paper.
We hope this will be useful when extending our work in the future by addressing these limitations.

On the theoretical side, as we already stated in \autoref{sec:exp}, the generative random walk model is not applicable to contextualised word embeddings obtained by a language model such as BERT.
As shown in \autoref{sec:context}, although the partition function for BERT demonstrates the self-normalising property when word embeddings are computed by averaging the context embeddings of a word across a corpus, the squared $\ell_2$ norm of these BERT-based word  does \emph{not} demonstrate a linear relationship with the logarithm of the frequency of that word in the corpus.
This has important consequences with regard to static sense embedding methods such as LMMS and ARES, which also use BERT to obtain sense representations from dictionaries such as the WordNet or sense-labelled corpora such as SemCor.
Although LMMS and ARES are full-coveraged state-of-the-art static sense embeddings, their $\ell_2$ norms does not satisfy the linear relationship, which we derived in this paper due to this reason as shown in \autoref{sec:context}.
An important future research direction would be to develop a random walk model for masked language models such as BERT, and analytically derive a relationship between the word embeddings and frequency of words.
We also exclude contextualised sense embedding methods such as SenseBERT~\cite{levine-etal-2020-sensebert} from our analysis due to the same reason.

On the empirical side, a limitation of our evaluation is that it is limited to the English language.
There are WSD and WiC benchmarks for other languages such as SemEval-13, SemEval-15, XL-WSD~\cite{pasini2021xl} and  WiC-XL~\cite{raganato-etal-2020-xl}, as well as multilingual sense embeddings such as ARES$_m$~\cite{scarlini2020more} and SensEmBERT~\cite{scarlini2020sensembert}.
% \textbf{@Jodie: Add some details here. Also cite some multilingual sense embedding here. We can then say we need to evaluate those sense embeddings for the linear relationship}.
Extending our evaluations to cover multilingual sense embeddings is deferred to future work.

\section{Ethical Considerations}
In this paper, we inspect the relationship between $\ell_2$ norm of static sense embedding and its frequency in the training corpus. 
We evaluate the effectiveness of $\ell_2$ norm of static sense embeddings on several experiments, i.e., MFS prediction, WiC and WSD tasks.
In particular, we did not annotate any datasets by ourselves in this work and used multiple corpora and benchmark datasets that have been collected, annotated and repeatedly used for evaluations in prior works.
To the best of our knowledge, no ethical issues have been reported concerning these datasets.
However, we note that it has been reported that pretrained sense embeddings encode various types of social biases such as gender and racial biases~\cite{zhou-etal-2022-sense}.
Our experiments show that including $\ell_2$ norm of sense embeddings to be an effective strategy for improving the performance in sense-related tasks such as WSD and WiC.
However, it remains an open question whether $\ell_2$ norm also encodes social biases, and if so how to mitigate those biases from affecting downstream tasks.

\bibliography{sense}

\begin{thebibliography}{27}
\expandafter\ifx\csname natexlab\endcsname\relax\def\natexlab#1{#1}\fi

\bibitem[{Andreas and Klein(2015)}]{andreas-klein:2015:NAACL-HLT}
Jacob Andreas and Dan Klein. 2015.
\newblock \href {https://aclanthology.org/N15-1027.pdf} {When and why are
  log-linear models self-normalizing?}
\newblock In \emph{Proceedings of the 2015 Conference of the North American
  Chapter of the Association for Computational Linguistics: Human Language
  Technologies}, pages 244--249, Denver, Colorado. Association for
  Computational Linguistics.

\bibitem[{Arora et~al.(2016)Arora, Li, Liang, Ma, and Risteski}]{Arora-word}
Sanjeev Arora, Yuanzhi Li, Yingyu Liang, Tengyu Ma, and Andrej Risteski. 2016.
\newblock \href {https://aclanthology.org/Q16-1028.pdf} {A latent variable
  model approach to pmi-based word embeddings}.
\newblock \emph{Transactions of the Association for Computational Linguistics},
  4:385--399.

\bibitem[{Arora et~al.(2018)Arora, Li, Liang, Ma, and Risteski}]{Arora:-sense}
Sanjeev Arora, Yuanzhi Li, Yingyu Liang, Tengyu Ma, and Andrej Risteski. 2018.
\newblock \href {https://aclanthology.org/Q18-1034.pdf} {Linear algebraic
  structure of word senses, with applications to polysemy}.
\newblock \emph{Transactions of the Association for Computational Linguistics},
  6:483--495.

\bibitem[{Bauerschmidt et~al.(2021)Bauerschmidt, Helmuth, and
  Swan}]{Bauerschmidt_2021}
Roland Bauerschmidt, Tyler Helmuth, and Andrew Swan. 2021.
\newblock \href {https://doi.org/10.1214/20-aihp1083} {The geometry of random
  walk isomorphism theorems}.
\newblock \emph{Annales de l{\textquotesingle}Institut Henri Poincar{\'{e}},
  Probabilit{\'{e}}s et Statistiques}, 57(1).

\bibitem[{Bhingardive et~al.(2015)Bhingardive, Singh, Rudramurthy, Redkar, and
  Bhattacharyya}]{bhingardive2015unsupervised}
Sudha Bhingardive, Dhirendra Singh, V~Rudramurthy, Hanumant~Harichandra Redkar,
  and Pushpak Bhattacharyya. 2015.
\newblock \href {https://aclanthology.org/N15-1132.pdf} {Unsupervised most
  frequent sense detection using word embeddings}.
\newblock In \emph{HLT-NAACL}.

\bibitem[{Devlin et~al.(2019)Devlin, Chang, Lee, and
  Toutanova}]{devlin2019bert}
Jacob Devlin, Ming-Wei Chang, Kenton Lee, and Kristina Toutanova. 2019.
\newblock \href {https://www.aclweb.org/anthology/N19-1423.pdf} {Bert:
  Pre-training of deep bidirectional transformers for language understanding}.
\newblock In \emph{Proceedings of the 2019 Conference of the North American
  Chapter of the Association for Computational Linguistics: Human Language
  Technologies, Volume 1 (Long and Short Papers)}, pages 4171--4186.

\bibitem[{Hashimoto et~al.(2016)Hashimoto, Alvarez-Melis, and
  Jaakkola}]{Hashimoto:TACL:2016}
Tatsunori Hashimoto, David Alvarez-Melis, and Tommi Jaakkola. 2016.
\newblock \href {https://aclanthology.org/Q16-1020.pdf} {Word embeddings as
  metric recovery in semantic spaces}.
\newblock \emph{Transactions of the Association for Computational Linguistics},
  4:273--286.

\bibitem[{Hauer et~al.(2019)Hauer, Luan, and Kondrak}]{hauer2019you}
Bradley Hauer, Yixing Luan, and Grzegorz Kondrak. 2019.
\newblock \href {https://ieeexplore.ieee.org/abstract/document/8665531} {You
  shall know the most frequent sense by the company it keeps}.
\newblock In \emph{2019 IEEE 13th International Conference on Semantic
  Computing (ICSC)}, pages 208--215. IEEE.

\bibitem[{Levine et~al.(2020)Levine, Lenz, Dagan, Ram, Padnos, Sharir,
  Shalev-Shwartz, Shashua, and Shoham}]{levine-etal-2020-sensebert}
Yoav Levine, Barak Lenz, Or~Dagan, Ori Ram, Dan Padnos, Or~Sharir, Shai
  Shalev-Shwartz, Amnon Shashua, and Yoav Shoham. 2020.
\newblock \href {https://doi.org/10.18653/v1/2020.acl-main.423} {{S}ense{BERT}:
  Driving some sense into {BERT}}.
\newblock In \emph{Proceedings of the 58th Annual Meeting of the Association
  for Computational Linguistics}, pages 4656--4667, Online. Association for
  Computational Linguistics.

\bibitem[{Loureiro and Jorge(2019{\natexlab{a}})}]{loureiro2019language}
Daniel Loureiro and Alipio Jorge. 2019{\natexlab{a}}.
\newblock \href {https://aclanthology.org/P19-1569} {Language modelling makes
  sense: Propagating representations through wordnet for full-coverage word
  sense disambiguation}.
\newblock In \emph{Proceedings of the 57th Annual Meeting of the Association
  for Computational Linguistics}, pages 5682--5691, Florence, Italy.

\bibitem[{Loureiro and Jorge(2019{\natexlab{b}})}]{loureiro2019liaad}
Daniel Loureiro and Alipio Jorge. 2019{\natexlab{b}}.
\newblock \href {https://aclanthology.org/W19-5801.pdf} {Liaad at semdeep-5
  challenge: Word-in-context (wic)}.
\newblock In \emph{Proceedings of the 5th Workshop on Semantic Deep Learning
  (SemDeep-5)}, pages 1--5.

\bibitem[{McCarthy et~al.(2007)McCarthy, Koeling, Weeds, and
  Carroll}]{McCarthy:2007}
Diana McCarthy, Rob Koeling, Julie Weeds, and John Carroll. 2007.
\newblock \href {https://aclanthology.org/J07-4005.pdf} {Unsupervised
  acquisition of predominant word senses}.
\newblock \emph{Computational Linguistics}, 33(4):553 -- 590.

\bibitem[{McCarthy et~al.(2004)McCarthy, Koeling, Weeds, and
  Carroll}]{mccarthy2004finding}
Diana McCarthy, Rob Koeling, Julie Weeds, and John~A Carroll. 2004.
\newblock \href {https://aclanthology.org/P04-1036.pdf} {Finding predominant
  word senses in untagged text}.
\newblock In \emph{Proceedings of the 42nd Annual Meeting of the Association
  for Computational Linguistics (ACL-04)}, pages 279--286.

\bibitem[{Mikolov et~al.(2013{\natexlab{a}})Mikolov, Chen, Corrado, and
  Dean}]{mikolov2013efficient}
Tomas Mikolov, Kai Chen, Greg Corrado, and Jeffrey Dean. 2013{\natexlab{a}}.
\newblock \href {https://arxiv.org/pdf/1301.3781.pdf} {Efficient estimation of
  word representations in vector space}.
\newblock \emph{arXiv preprint arXiv:1301.3781}.

\bibitem[{Mikolov et~al.(2013{\natexlab{b}})Mikolov, Sutskever, Chen, Corrado,
  and Dean}]{mikolov2013distributed}
Tomas Mikolov, Ilya Sutskever, Kai Chen, Greg~S Corrado, and Jeff Dean.
  2013{\natexlab{b}}.
\newblock \href
  {https://proceedings.neurips.cc/paper/2013/file/9aa42b31882ec039965f3c4923ce901b-Paper.pdf}
  {Distributed representations of words and phrases and their
  compositionality}.
\newblock In \emph{Advances in NIPS}, pages 3111--3119.

\bibitem[{Miller(1995)}]{miller1995wordnet}
George~A Miller. 1995.
\newblock \href {https://dl.acm.org/doi/abs/10.1145/219717.219748} {Wordnet: a
  lexical database for english}.
\newblock \emph{Communications of the ACM}, 38(11):39--41.

\bibitem[{Miller et~al.(1993)Miller, Leacock, Tengi, and
  Bunker}]{miller1993semantic}
George~A Miller, Claudia Leacock, Randee Tengi, and Ross~T Bunker. 1993.
\newblock \href {https://aclanthology.org/H93-1061.pdf} {A semantic
  concordance}.
\newblock In \emph{Human Language Technology: Proceedings of a Workshop Held at
  Plainsboro, New Jersey, March 21-24, 1993}.

\bibitem[{Mu and Viswanath(2018)}]{mu2018allbutthetop}
Jiaqi Mu and Pramod Viswanath. 2018.
\newblock \href {https://arxiv.org/pdf/1702.01417.pdf} {All-but-the-top: Simple
  and effective postprocessing for word representations}.
\newblock In \emph{International Conference on Learning Representations}.

\bibitem[{Pasini and Navigli(2018)}]{pasini2018two}
Tommaso Pasini and Roberto Navigli. 2018.
\newblock \href {https://ojs.aaai.org/index.php/AAAI/article/view/11961} {Two
  knowledge-based methods for high-performance sense distribution learning}.
\newblock In \emph{Proceedings of the AAAI Conference on Artificial
  Intelligence}, volume~32.

\bibitem[{Pasini et~al.(2021)Pasini, Raganato, and Navigli}]{pasini2021xl}
Tommaso Pasini, Alessandro Raganato, and Roberto Navigli. 2021.
\newblock \href {https://ojs.aaai.org/index.php/AAAI/article/view/17609}
  {X{L}-{WSD}: An extra-large and cross-lingual evaluation framework for word
  sense disambiguation}.
\newblock In \emph{Proceedings of the AAAI Conference on Artificial
  Intelligence}, volume~35, pages 13648--13656.

\bibitem[{Pennington et~al.(2014)Pennington, Socher, and
  Manning}]{pennington2014glove}
Jeffrey Pennington, Richard Socher, and Christopher~D Manning. 2014.
\newblock \href {https://aclanthology.org/D14-1162.pdf} {Glo{V}e: Global
  vectors for word representation}.
\newblock In \emph{Proceedings of the 2014 conference on empirical methods in
  natural language processing (EMNLP)}, pages 1532--1543.

\bibitem[{Pilehvar and Camacho-Collados(2019)}]{Pilehvar:2019}
Mohammad~Taher Pilehvar and Jose Camacho-Collados. 2019.
\newblock \href {https://doi.org/10.18653/v1/N19-1128} {{W}i{C}: the
  word-in-context dataset for evaluating context-sensitive meaning
  representations}.
\newblock In \emph{Proceedings of the 2019 Conference of the North {A}merican
  Chapter of the Association for Computational Linguistics: Human Language
  Technologies, Volume 1 (Long and Short Papers)}, pages 1267--1273,
  Minneapolis, Minnesota. Association for Computational Linguistics.

\bibitem[{Raganato et~al.(2017)Raganato, Camacho-Collados, and
  Navigli}]{raganato2017word}
Alessandro Raganato, Jose Camacho-Collados, and Roberto Navigli. 2017.
\newblock \href {https://aclanthology.org/E17-1010.pdf} {Word sense
  disambiguation: A unified evaluation framework and empirical comparison}.
\newblock In \emph{Proceedings of the 15th Conference of the European Chapter
  of the Association for Computational Linguistics: Volume 1, Long Papers},
  pages 99--110.

\bibitem[{Raganato et~al.(2020)Raganato, Pasini, Camacho-Collados, and
  Pilehvar}]{raganato-etal-2020-xl}
Alessandro Raganato, Tommaso Pasini, Jose Camacho-Collados, and Mohammad~Taher
  Pilehvar. 2020.
\newblock \href {https://doi.org/10.18653/v1/2020.emnlp-main.584}
  {{XL}-{W}i{C}: A multilingual benchmark for evaluating semantic
  contextualization}.
\newblock In \emph{Proceedings of the 2020 Conference on Empirical Methods in
  Natural Language Processing (EMNLP)}, pages 7193--7206, Online. Association
  for Computational Linguistics.

\bibitem[{Scarlini et~al.(2020{\natexlab{a}})Scarlini, Pasini, and
  Navigli}]{scarlini2020sensembert}
Bianca Scarlini, Tommaso Pasini, and Roberto Navigli. 2020{\natexlab{a}}.
\newblock \href {https://ojs.aaai.org//index.php/AAAI/article/view/6402}
  {Sens{E}m{BERT}: Context-enhanced sense embeddings for multilingual word
  sense disambiguation}.
\newblock In \emph{Proceedings of the AAAI Conference on Artificial
  Intelligence}, volume~34, pages 8758--8765.

\bibitem[{Scarlini et~al.(2020{\natexlab{b}})Scarlini, Pasini, and
  Navigli}]{scarlini2020more}
Bianca Scarlini, Tommaso Pasini, and Roberto Navigli. 2020{\natexlab{b}}.
\newblock \href {https://aclanthology.org/2020.emnlp-main.285/} {With more
  contexts comes better performance: Contextualized sense embeddings for
  all-round word sense disambiguation}.
\newblock In \emph{Proceedings of the 2020 Conference on Empirical Methods in
  Natural Language Processing (EMNLP)}, pages 3528--3539, Online.

\bibitem[{Zhou et~al.(2022)Zhou, Kaneko, and Bollegala}]{zhou-etal-2022-sense}
Yi~Zhou, Masahiro Kaneko, and Danushka Bollegala. 2022.
\newblock \href {https://aclanthology.org/2022.acl-long.135.pdf} {Sense
  embeddings are also biased {--} evaluating social biases in static and
  contextualised sense embeddings}.
\newblock In \emph{Proceedings of the 60th Annual Meeting of the Association
  for Computational Linguistics (Volume 1: Long Papers)}, pages 1924--1935,
  Dublin, Ireland. Association for Computational Linguistics.

\end{thebibliography}
\bibliographystyle{acl_natbib}

\appendix
\section*{Appendix}

\section{Experimental Setup}
In this section, we report the setup of experiments we conduct to evaluate the effectiveness of $\ell_2$ norm of static sense embeddings.

\subsection{Training GloVe-sense and SGNS-sense}
We train our GloVe-sense and SGNS-sense on SemCor training data.
Specifically, for each target word $w$ in a context $c$, we train a vector and assign the annotated sense label to it. 
For GloVe-sense, we use its Python-based implementation.\footnote{\url{https://github.com/maciejkula/glove-python}}
We set the co-occurrence window to $10$ tokens, number of dimensions to $300$ and the initial learning rate to $0.05$ for the vanilla stochastic gradient descent.
We train the embeddings for $30$ epochs with $2$ parallel threads.
To train SGNS-sense, we use the Word2Vec module from gensim.models.\footnote{\url{https://radimrehurek.com/gensim/models/word2vec.html}}
We set the min\_count to $1$ and the dimensionality of the embeddings to $300$, and the remainder of the hyperparameters remain at their default values.

\subsection{Predicting the Most Frequent Sense}
We conduct our MFS experiment on SemCor. 
Given a target word $w$ in a context $c$, we first select a set of candidate senses based on $w$'s lemma and PoS. 
Then we compute the $\ell_2$ norm of the static sense embedding for each sense in the candidate set.
Finally, we take the sense with the maximum $\ell_2$ norm score as the predicted MFS for $w$.
Then we compare our prediction with the MFS of $w$ according to the sense occurrence in SemCor, and compute the accuracy scores. 

\subsection{Word Sense Disambiguation}
\label{appendix:sec:WSD}
We consider the Word Sense Disambiguation task as a binary classification problem and train a Logistic Regression binary classifier on SemCor. 
To evaluate the baselines, i.e., LMMS (LMMS SP-WSD: sensekeys\footnote{\url{https://github.com/danlou/LMMS}}) and ARES on WSD, given a word $w$ in a sentence $c$, we first compute its contextualised embedding using BERT (bert-large-cased) model by averaging the last four layers, denoted by $\vec{t}(w, c)$.
We then compute the cosine similarity between $\vec{t}(w, c)$ and the sense embedding $\vec{s}(w)$ corresponding to the senses of $w$ based on WordNet as a feature. 
We use the binary logistic regression classifier implemented in sklearn with the default parameters.
For our proposed method, we simply append the $\ell_2$ norm of static sense embedding of $w$ as an additional feature. 
To avoid any discrepancies in the scoring methodology, we use the English all-words WSD framework~\cite{raganato2017word} and its offical scoring scripts. 

% \begin{figure}[t]
% \centering
% \includegraphics[width=0.48\textwidth]{figures/SGNS-partition-function-histogram-final.png}
% \caption{Histogram of partition function for 1,000 random vectors $\vec{c}$ for SGNS-sense. The x-axis is normalised by the mean of the values.}
% \label{fig:sgns-partition}
% \end{figure}

% \begin{figure}[t]
% \centering
% \includegraphics[width=0.48\textwidth]{figures/l2_embeddings-frequency-SGNS-sense.png}
% \caption{A linear relationship can be seen between the squared $\ell_2$ norms of SGNS-sense embeddings and the logarithms of the sense frequencies. Each dot in the plot corresponds to a sense, where x-axis is the natural logarithm of the sense frequency, and y-axis is the squared $\ell_2$ norm of the word embedding. The Pearson correlation coefficient between the two variables is $0.440$.}
% \label{fig:sgns-corr}
% \end{figure}

\subsection{Predicting Word Sense in Context}
We train a binary logistic regression classifier\footnote{We use the default parameters in \url{scikit-learn.org/stable/modules/generated/sklearn.linear_model.LogisticRegression.html}.} on the WiC training set. 
Following the work from~\newcite{loureiro2019liaad}, we compute four similarities between sense and contextualised embeddings, and consider those as features.
Specifically, given a target word $w$ in two contexts $c_1$ and $c_2$, similar to~\autoref{sec:WSD}, we first determine the sense-specific embeddings for $w$ in $c_1$ and $c_2$, denoted by $\vec{s}_{1}(w)$ and $\vec{s}_{2}(w)$.
Then we use the cosine similarities between the two vectors in the following four pairs as features, requiring no expensive fine-tuning procedure: ($\vec{s}_{1}(w)$, $\vec{s}_{2}(w)$), ($\vec{t}(w,c_{1})$, $\vec{t}(w,c_{2})$), ($\vec{s}_{1}(w)$, $\vec{t}(w,c_{1})$), ($\vec{s}_{2}(w)$, , $\vec{t}(w,c_{2})$).
Contextualised embeddings are not $\ell_2$ normalised in this experiment.
Here again, similar to the WSD settings described above, with respect to our proposed method, we simply append the $\ell_2$ norm of the static sense embedding of $w$ as the fifth feature. 

\begin{figure}[t]
\centering
\includegraphics[width=0.48\textwidth,height=0.35\textwidth]{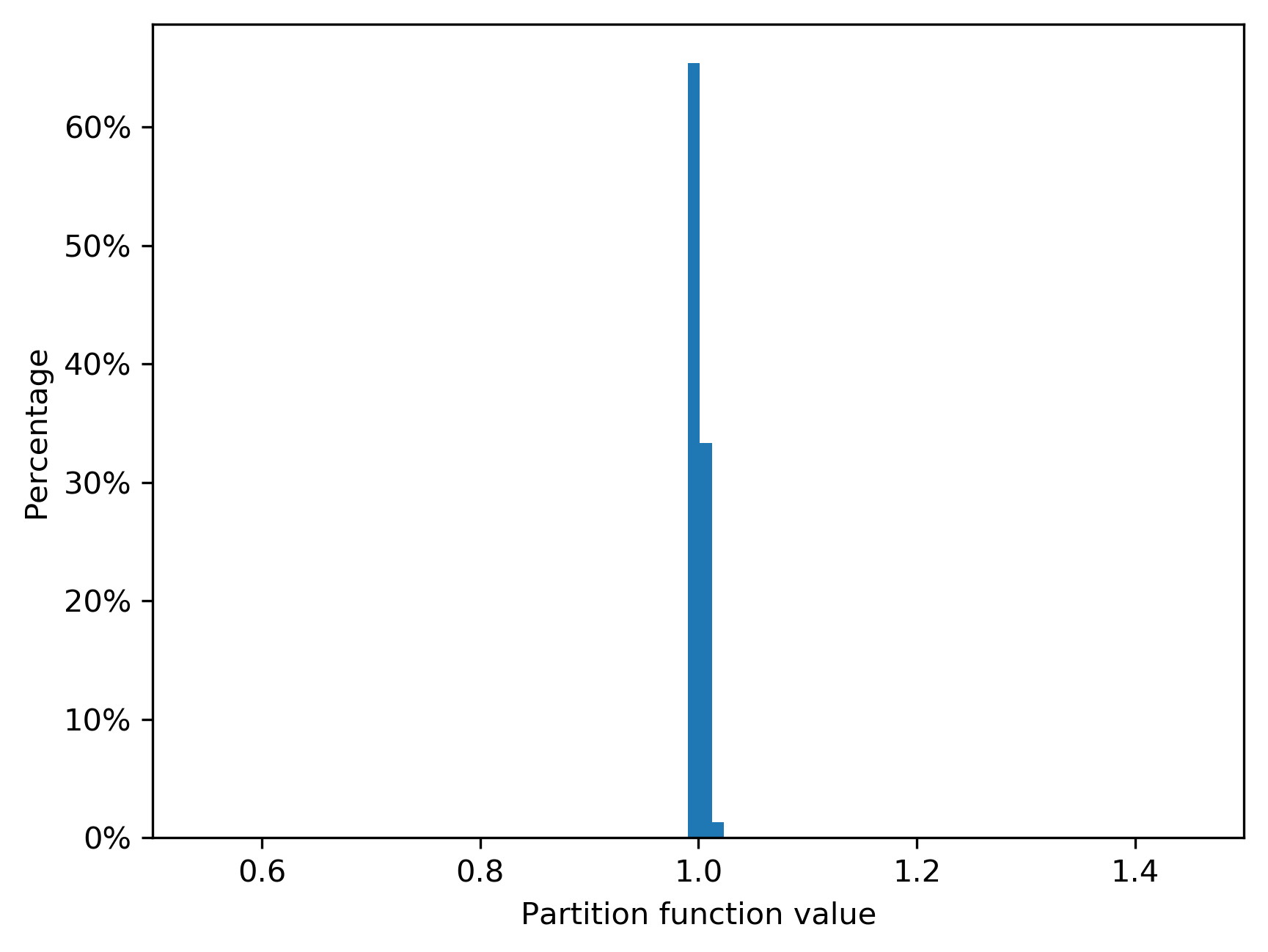}
\caption{Histogram of the partition function for 1,000 random vectors $c$ for BERT-word. The x-axis is normalised by the mean of the values.}
\label{fig:bert-partition}
\end{figure}

\begin{figure}[t]
\centering
\includegraphics[width=0.48\textwidth,height=0.35\textwidth]{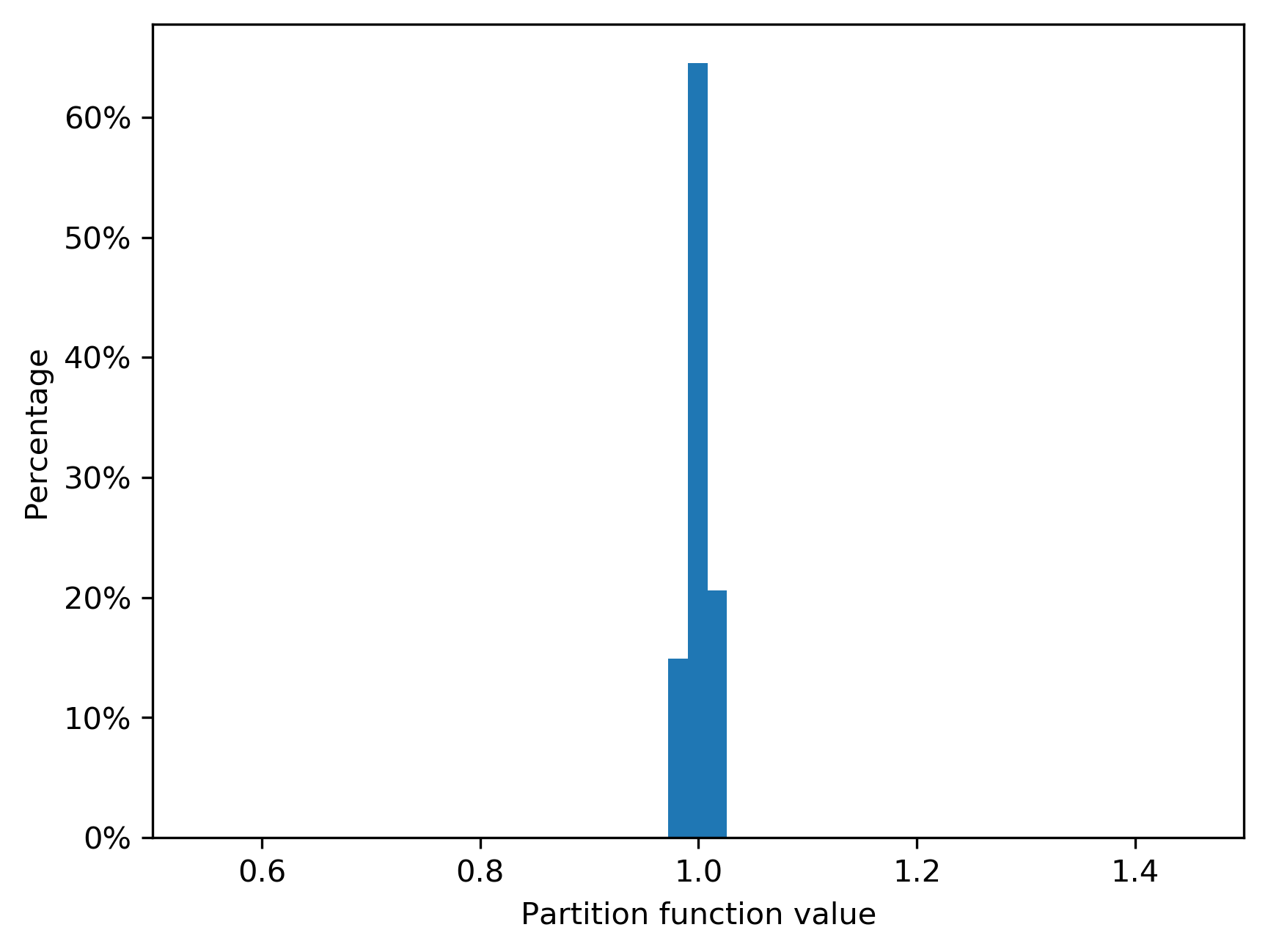}
\caption{Histogram of the partition function for 1,000 random vectors $c$ for LMMS. The x-axis is normalised by the mean of the values.}
\label{fig:lmms-partition}
\end{figure}

\section{Static Sense Embeddings from Contextualised Word Embeddings}
\label{sec:context}

\begin{figure}[t]
\centering
\includegraphics[width=0.48\textwidth,height=0.35\textwidth]{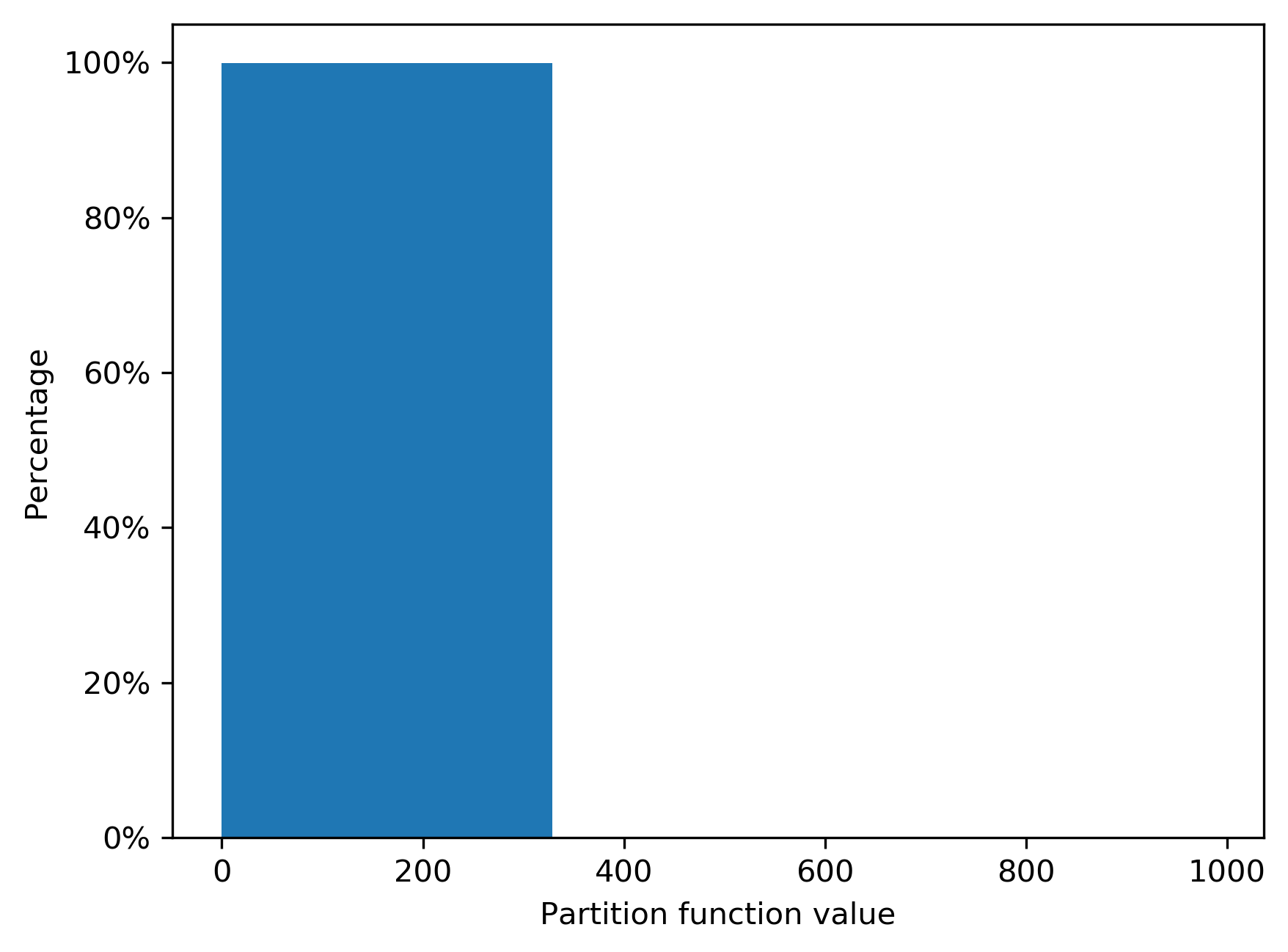}
\caption{Histogram of partition function for 1,000 random vectors c for LMMS$_{sc}$. The x-axis is normalised by the mean of the values.}
\label{fig:lmms-sc-partition}
\end{figure}

\begin{figure}[t]
\centering
\includegraphics[width=0.48\textwidth,height=0.35\textwidth]{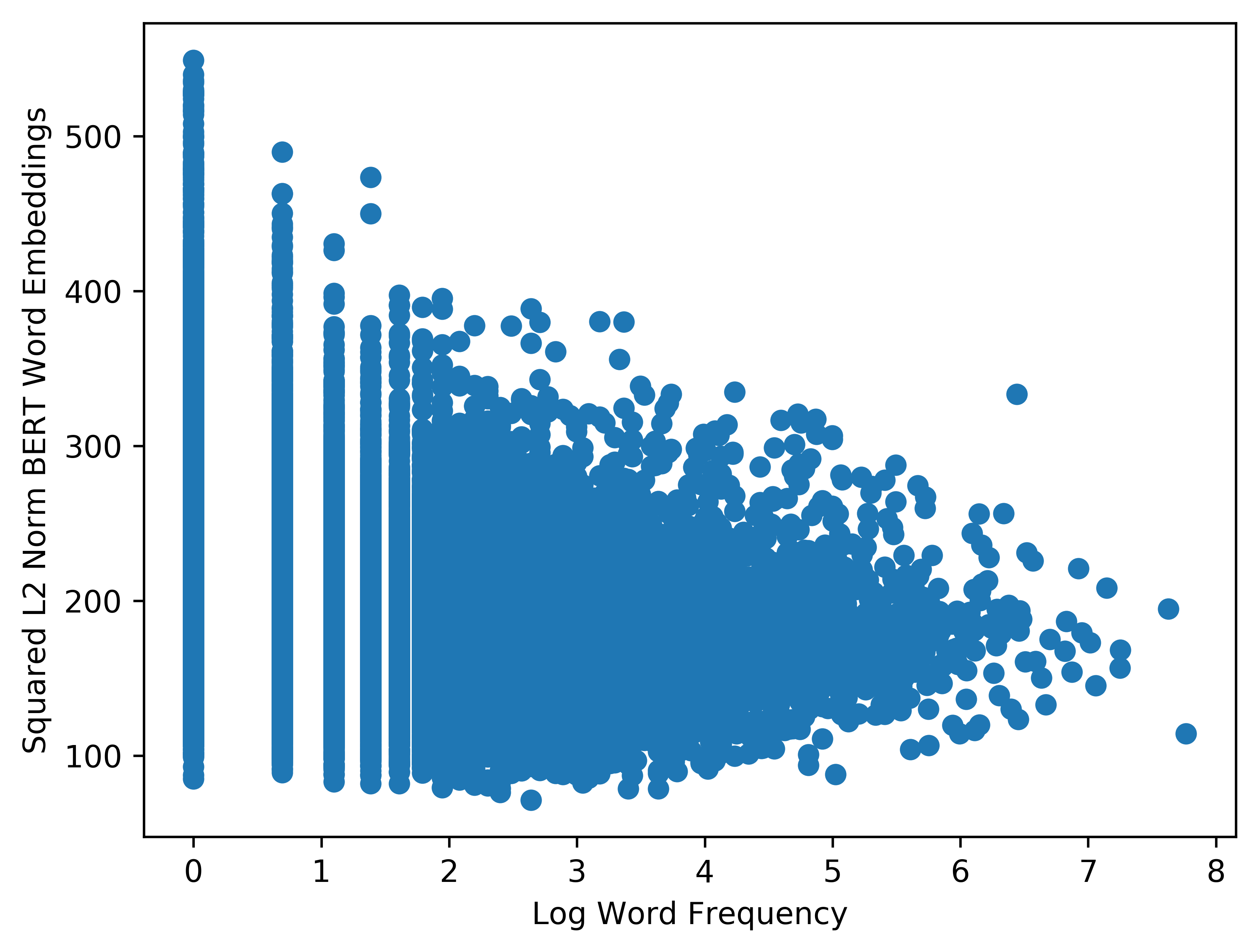}
\caption{The linear relationship between the squared $\ell_2$ norms of BERT-word embeddings and the logarithms of the sense frequencies. Each dot in the plot corresponds to a word, where x-axis is the natural logarithm of the sense frequency, and y-axis is the squared $\ell_2$ norm of the word embedding. The Pearson correlation coefficient between the two is $-0.316$.}
\label{fig:bert-corr}
\end{figure}

We investigate whether the self-normalising and linearity properties hold for contextualised embeddings obtained from language models.
For this purpose, we compute the static word embeddings for the words appearing in SemCor using contextualised embeddings learnt by BERT.
Specifically, we compute the average over the contextualised BERT embeddings for all of the occurrences of a word in SemCor, and consider it as the static (i.e. context-independent) BERT embedding for that word. 
To distinguish the contextualised embeddings learnt from BERT, we name the static BERT embeddings as BERT-static in the remainder of this paper.

\begin{figure}[t]
\centering
\includegraphics[width=0.48\textwidth,height=0.35\textwidth]{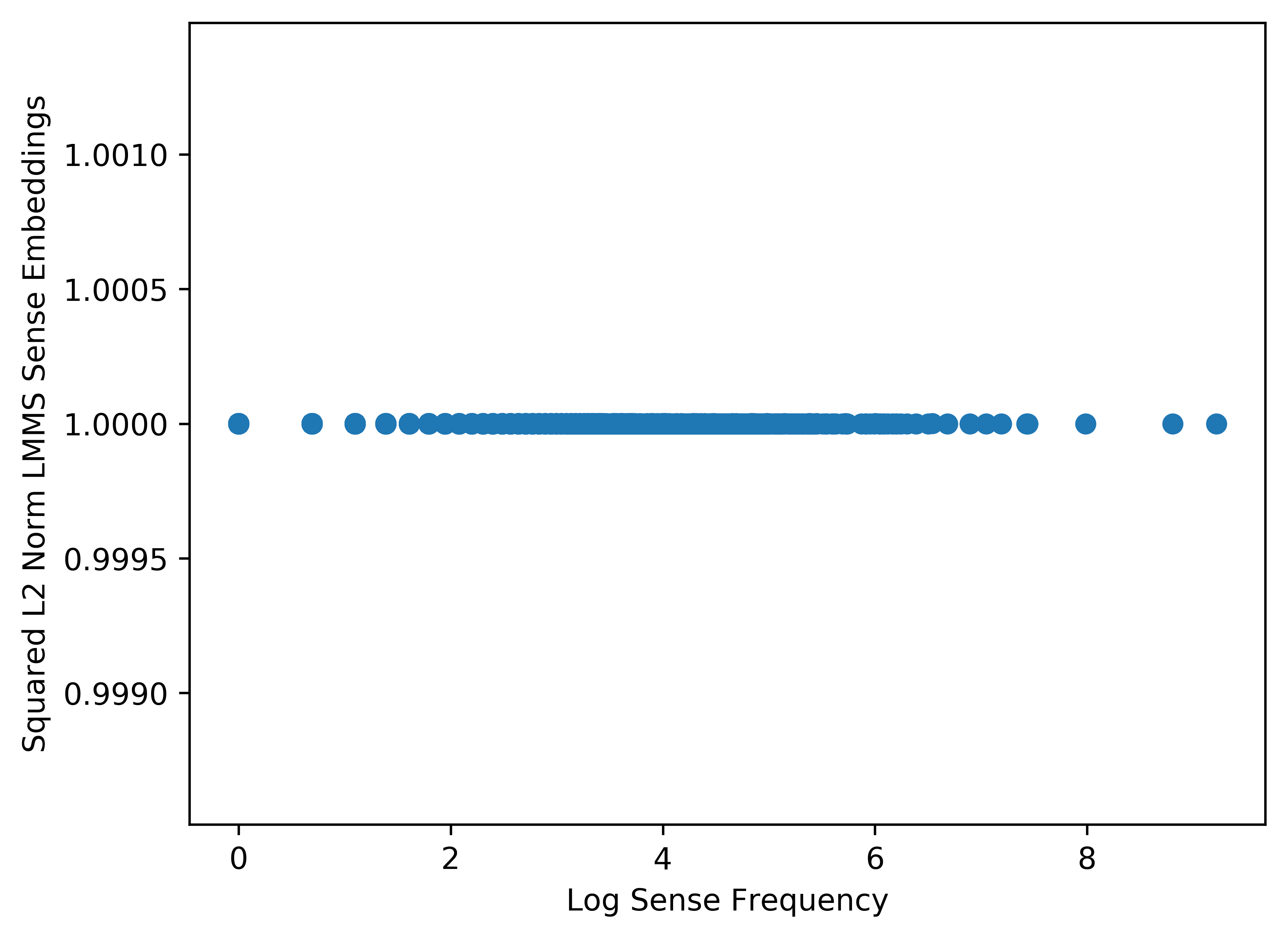}
\caption{The linear relationship between the squared $\ell_2$ norms of LMMS embeddings and the logarithms of the sense frequencies. Each dot in the plot corresponds to a word, where x-axis is the natural logarithm of the sense frequency, and y-axis is the squared $\ell_2$ norm of the word embedding. The Pearson correlation coefficient between the two is $-0.005$.}
\label{fig:lmms-corr}
\end{figure}

Recall that LMMS uses BERT to compute sense embeddings from SemCor and WordNet's glosses.
Therefore, if BERT-static satisfies the self-normalising and linearity properties described in \autoref{sec:theory}, LMMS embedding must satisfy these properties as well. 
In addition, we take the first step of LMMS training procedure from the work of~\newcite{loureiro2019language}\footnote{\url{https://github.com/danlou/LMMS}} and
train static sense embeddings only on SemCor data without normalising the learnt sense embeddings (doing so would remove $\ell_2$ norm related information from the sense embeddings).
To differentiate this version of LMMS embeddings from the full-coverage LMMS embeddings, we refer to it as LMMS$_{sc}$ (here, sc stands for SemCor).
We then test if the self-normalising and linearity properties hold for BERT-static, LMMS and LMMS$_{sc}$.

\autoref{fig:bert-partition}, \ref{fig:lmms-partition} and \ref{fig:lmms-sc-partition} show the histogram of partition functions for BERT-word, LMMS and LMMS$_{sc}$, respectively.
We observe that the histograms of both BERT-static and LMMS are centred around mean, while LMMS$_{sc}$ is not.
This shows that LMMS$_{sc}$ does not satisfy self normalising, while BERT-static and LMMS do.

\begin{figure}[H]
\centering
\includegraphics[width=0.48\textwidth,height=0.35\textwidth]{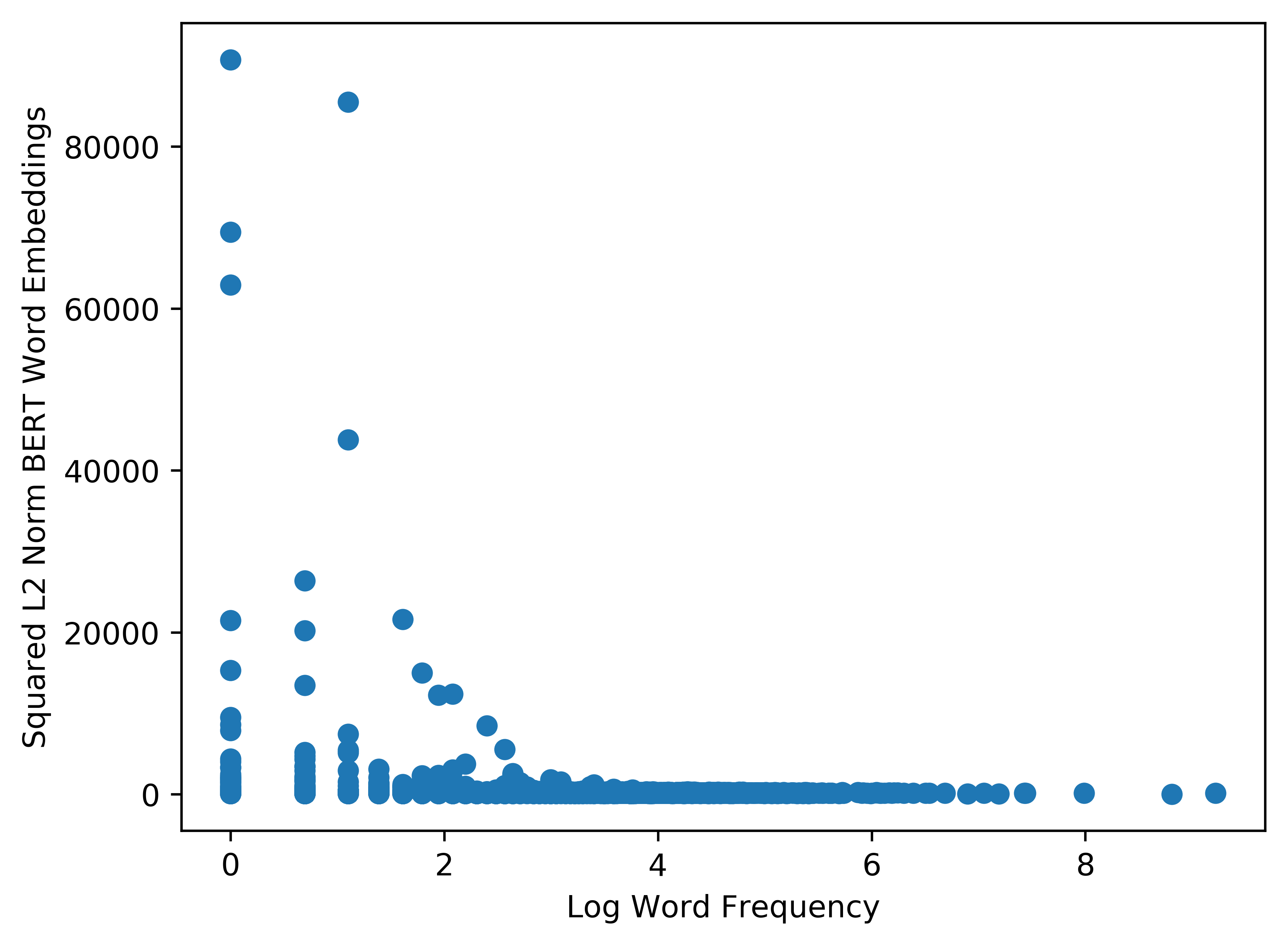}
\caption{The linear relationship between the squared $\ell_2$ norms of LMMS$_{sc}$ embeddings and the logarithms of the sense frequencies. Each dot in the plot corresponds to a word, where x-axis is the natural logarithm of the sense frequency, and y-axis is the squared $\ell_2$ norm of the word embedding. The Pearson correlation coefficient between the two is $-0.010$.}
\label{fig:lmms-sc-corr}
\end{figure}

\autoref{fig:bert-corr}, \ref{fig:lmms-corr} and \ref{fig:lmms-sc-corr} show the correlation between squared $\ell_2$ norms of the word/sense embeddings and the logarithms of sense/word frequencies for BERT-static, LMMS and LMMS$_{sc}$, respectively.
From the figures we see that none shows a linear relationship.
This indicates that sense frequency related information is not encoded in the $\ell_2$ norm of LMMS (or BERT) embeddings.
\end{document}